\documentclass{article}

    \PassOptionsToPackage{numbers, compress}{natbib}

\usepackage[final]{neurips_2021_preregistration}

\usepackage[utf8]{inputenc} 
\usepackage[T1]{fontenc}    
\usepackage{hyperref}       
\usepackage{url}            
\usepackage{booktabs}       
\usepackage{amsfonts}       
\usepackage{amsmath}        
\usepackage{nicefrac}       
\usepackage{microtype}      
\usepackage{xcolor}         
\usepackage{amssymb}
\usepackage{graphicx}
\usepackage{enumitem}

\title{Open-Set Multi-Source Multi-Target \\ Domain Adaptation}

\author{Rohit Lal \textsuperscript{1}, Arihant Gaur \textsuperscript{2}, Aadhithya Iyer \textsuperscript{2}, Muhammed Abdullah Shaikh\textsuperscript{2},\\ \textbf{Ritik Agrawal \textsuperscript{2}}, \textbf{Shital Chiddarwar\textsuperscript{2}}
\\
\textsuperscript{1}Indian Institute of Science (IISc), Bangalore\\ \textsuperscript{2} Visvesvaraya National Institute of Technology (VNIT), Nagpur}

\begin{document}

\maketitle
\begin{abstract}

Single-Source Single-Target Domain Adaptation (\textit{\texttt{1}S\texttt{1}T}) aims to bridge the gap between a labelled source domain and an unlabelled target domain. Despite \textit{\texttt{1}S\texttt{1}T} being a well-researched topic, they are typically not deployed to the real world. Methods like Multi-Source Domain Adaptation and Multi-Target Domain Adaptation have evolved to model real-world problems but still do not generalise well. The fact that most of these methods assume a common label-set between source and target is very restrictive. Recent Open-Set Domain Adaptation methods handle unknown target labels but fail to generalise in multiple domains. To overcome these difficulties, first, we propose a novel generic domain adaptation (DA) setting named Open-Set Multi-Source Multi-Target Domain Adaptation (\textit{OS-nSmT}), with $n$ and $m$ being number of source and target domains respectively. Next, we propose a graph attention based framework named DEGAA which can capture information from multiple source and target domains without knowing the exact label-set of the target. We argue that our method, though offered for multiple sources and multiple targets, can also be agnostic to various other DA settings. To check the robustness and versatility of DEGAA, we put forward ample experiments and ablation studies. 




\end{abstract}

\section{Introduction} \label{introduction}

Recent advances in deep learning have achieved state-of-the-art (SOTA) results in tasks like image classification \citep{Krizhevsky_imagenetclassification}, object detection and segmentation \citep{DBLP:journals/corr/GirshickDDM13}. Most of these methods require a lot of labelled data and assumes that all data comes from identical distributions (i.i.d.) that is, training data distribution is the same as test data distribution. This assumption rarely holds for real-world cases. In Unsupervised Domain Adaptation (UDA), we have labelled source data and unlabelled target data. The source and target data here belong to different distributions. The goal of UDA is to overcome this distributional shift and perform well on both labelled source and unlabelled target data. 

Even though closed-set Single-Source Single-Target (\textit{\texttt{1}S\texttt{1}T}), Multi-Source Single Target (\textit{nS\texttt{1}T}), Single Source Multi-Target (\textit{\texttt{1}SmT}) and Multi-Source Multi-Target (\textit{nSmT}) settings show promising results, these methods share a common label set between the source and target domains (closed-set labels). This assumption may not hold for models deployed in the wild as the model may encounter unknown classes as well. The aforementioned restrictions motivated us to look for a new setting that can model real-world problems more closely and is scalable to a vast range of problems. 
In this paper, we propose a novel Domain Adaptation (DA) setting named Open-Set Multi-Source Multi-Target Unsupervised Domain Adaptation (\textit{OS-nSmT}) that closely resembles the real-world scenarios. In \textit{OS-nSmT}, we have access to multiple labelled source domain dataset with the same label set and multiple unlabelled target domain datasets. Here, we consider the source label set as a subset of the target label set. This is a very difficult problem to solve as one has to overcome both the category and domain gap. 
Graphs Neural Networks (GNN) based approaches \cite{roy2021curriculum, 9204804, ma2019gcan, luo2020progressive} have gained traction in this field due to their ability to find relations in unstructured data. It also helps to represent the domains in a unified subspace. We use GNN with attentional aggregation to find optimal relationships between the samples and perform semantic transfer accordingly. We propose an algorithm named \textbf{D}omain \textbf{E}mbedding based \textbf{G}raph \textbf{A}ttentional \textbf{A}ggregation (DEGAA) to solve this important yet unexplored setting. This algorithm finds potential application in autonomous systems, where rather than adapting the model for one country at a time, we can incorporate data for multiple countries. This work can be potentially extended in the field of domain generalization in which we have multiple unseen target domains. Other applications might include face recognition, cross-lingual learning in Natural Language Processing, sentiment classification, etc. To summarise, the work proposes the following contributions:

\begin{itemize}[noitemsep, nolistsep, leftmargin=*]
    \item Introduction of a novel DA setting (\textit{OS-nSmT}) which is close to modelling real world applications.
    \item Development of DEGAA: A Graph Attentional Aggregation based algorithm that can work with all kind of DA setting (\textit{\texttt{1}S\texttt{1}T}, \textit{\texttt{1}SmT}, \textit{nSmT}, etc) with open-set labels.
\end{itemize}



\vspace{-0.1cm}
\section{Related works} \label{related-works}
\vspace{-0.2cm}
\textbf{(a) Closed set domain adaptation (CSDA).} CSDA is a setting with an assumption of shared support between source and target domains. A common method is to minimize the domain discrepancy using the Moment Matching technique which utilizes various metrics like MMD \cite{Long2015LearningTF}, CMD \cite{Zellinger2017CentralMD} and other variations \cite{6751479,6136518, long2016unsupervised, 10.5555/3305890.3305909, Saito_2018_CVPR, Yan2017MindTC, Na_2021_CVPR}. Alternatively, to reduce the domain shift, adversarial networks have been used to maximize the confusion between source and target domains \cite{pmlr-v37-ganin15, JMLR:v17:15-239, Tzeng_2017_CVPR, Kang_2018_ECCV, NEURIPS2018_ab88b157}. With the advancement in image synthesis using GAN \cite{goodfellow2014generative}, various generative model-based methods \cite{pmlr-v80-hoffman18a, Bousmalis_2017_CVPR, Hu_2018_CVPR, Zhu2017UnpairedIT, Sankaranarayanan_2018_CVPR, Volpi2018AdversarialFA, Yang2020BiDirectionalGF} have been proposed.
Though researchers have achieved good accuracy in \textit{\texttt{1}S\texttt{1}T}, it can't leverage the huge amount of data that is readily available across multiple different domains. Recent progress in DA settings like \textit{nS\texttt{1}T} \cite{Cui2020MultiSourceAF, NEURIPS2018_717d8b3d, peng2019moment, xu2018deep} and \textit{\texttt{1}SmT} \cite{Isobe_2021_CVPR, peng2019domain, Chen_2019_CVPR, nguyen2021unsupervised} tries to take advantage of multiple sources/targets.
\textit{nSmT} \cite{9242273, jin2020minimum} largely remains unexplored due to its increased complexity. While few methods \cite{jin2020minimum} in \textit{nSmT} handle multiple DA settings, they still can't handle unseen target labels. 

\textbf{(b) Open set domain adaptation (OSDA).} In OSDA setting there aren't any assumptions that source and target should have common labels.
Prior work \cite{Busto_2017_ICCV, Saito_2018_ECCV, 7780542, ijcai2020-352} trains feature generator such that the private class samples from target  domain are assigned as  "\textit{unknown}", \cite{Liu_2019_CVPR} uses coarse-to-fine refining of samples, while other methods \cite{ge2017generative, 8461700} harness the capability of GAN for generating more "unknown" samples for training. A survey on OSDA \cite{geng2020recent} shows that prior work in open set setting is mostly for \textit{\texttt{1}S\texttt{1}T} scenario. Rakshit \textit{et al}.\cite{Rakshit2020MultisourceOD} introduced the paradigm of \textit{nS\texttt{1}T} in Open Set scenario and uses adverserial training strategy, while other cases are yet to be explored. Our work proposes a framework for a new setting i.e \textit{OS-nSmT} which can further be utilized for other scenarios as well. 

\textbf{(c) Graph neural networks (GNN).} GNN are essentially neural network models, directly applied to data structured as a graph, capturing relationships between the nodes through the propagation of messages from nearby nodes \cite{gori2005new, wu2020comprehensive}. Recent methods \cite{9204804,ma2019gcan, luo2020progressive} involving GNNs have gained popularity owing to their transductive ability for semantic propagation of related samples among multiple domains.
Roy \textit{et al.} \cite{roy2021curriculum} proposes using graph convolution networks in \textit{\texttt{1}SmT} setting, along with co-curriculum teaching to handle noisy pseudo-labels. In our novel setting, we propose the use of attentional graph neural network \cite{sarlin2020superglue} by passing messages between the source and target domains to aggregate them in a unified space.

\textbf{(d) Domain generalization.} Domain generalization involves training on multiple observed domains, while expecting it to work well on unseen domains. In contrast to domain adaptation, there is no bridge network to distill the knowledge from source to target domain. We leverage the use of domain embeddings \cite{dubey2021adaptive}, that were originally used for domain generalization, by taking both the input and domain into account to generate pseudo - labels for target domain. A recent method \cite{wang2021tent} used entropy minimization technique for modulating features to learn domain-invariant representations. Other methods include using adversarial training \cite{sun2016deep, li2018domain, li2018deep} inspired from \cite{JMLR:v17:15-239},  meta learning \cite{li2018learning}, domain mixup \cite{xu2020adversarial, yan2020improve, wang2020heterogeneous}, distributionally robust optimization \cite{sagawa2019distributionally}, causal matching \cite{mahajan2021domain} and invariant risk minimization \cite{arjovsky2019invariant}. 


\section{Methodology} \label{methodology}
\paragraph{Notations} We define $F(\mathbf{x};\theta): \mathbb{R}^{3 \times w \times h} \to \mathbb{R}^d$ as the backbone network, parameterized by weights $\theta$ and input $\mathbf{x}$. The domain embedding network is represented as $G(\mathbf{x};\psi): \mathbb{R}^{3 \times w \times h} \to \mathbb{R}^{d}$, that yields the vector corresponding to each domain. Being an \textit{nSmT} problem, there are $n$ source domains ($S = \cup_{i = 1}^{n}S_i$) and $m$ target domains ($T = \cup_{j = 1}^{m}T_j$). They are collectively defined in the domain set $D = S \cup T$. 

In the following section, we will discuss the full architecture in detail (Fig. \ref{fig:arch}). For algorithm, please refer to the supplementary material. 

\subsection{Domain embedding} \label{domain-embedding}
We follow an approach similar to  \cite{dubey2021adaptive} for obtaining domain embeddings. We propose to train the domain embedding network in a prototypical fashion \cite{snell2017prototypical}, over a subset of points from each domain, divided into support and query points. They are sampled from the union of source and target domains. The support points yield an approximate embedding for each domain. Using this embedding, the probability of occurrence of query points in that domain is estimated, whose negative log-likelihood is minimized in an unsupervised fashion. In other words, the query is used to minimize the distance from the embeddings obtained using support points. During its inference, the Kernel Mean Embedding (KME)  $\pmb{\mu}$ \cite{muandet2016kernel} for domain is determined as,
\begin{equation}
    \pmb{\mu}(D_i)=\frac{1}{N(D_i)}\sum_{\mathbf{x}\in D_i}G(\mathbf{x}; \psi)
\end{equation}
Where $N(D_i)$ represents number of points in domain $D_i$. 
We store the KME of all the domains as a dictionary, to obtain a domain embedding prototype, $d_{e}$.
We propose that the domain embedding acts as additional information. This enables effective domain adaptation to unseen domains as it is possible to learn domain embedding without having access to labels. The new feature vector generated by concatenating image representation and domain embedding, can be used to project source and target image to a common feature space.

\subsection{Stage 1: Warm-Up Stage} \label{warmup}







\begin{figure}
    \centering
    \includegraphics[scale=0.69]{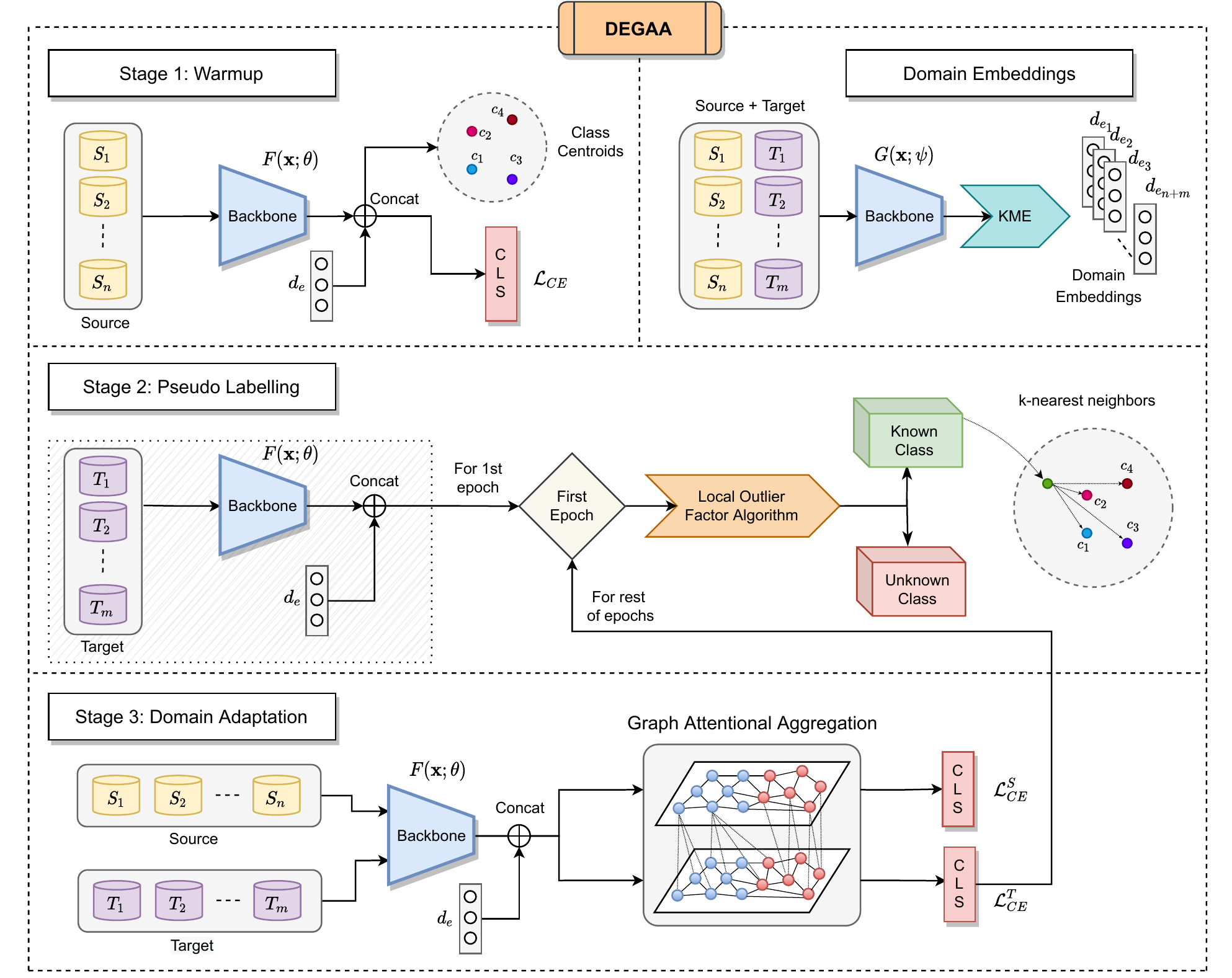}
    \caption{\textbf{DEGAA Architecture}. In stage 1, we do supervised source training by concatenating image features with domain embeddings. This model is initially used in stage 2 for generating pseudo-labels. In stage 3, we utilise the pseudo-labels generated from stage 2 for domain adaptation. For the rest of the epochs, we toggle between stage 2 and stage 3 for iterative training.}
    \label{fig:arch}
\end{figure}

We propose to train the feature extractor, in a supervised fashion, with the labels from source domains. The extracted feature, concatenated with the domain embedding from Section \ref{domain-embedding} is used to calculate the corresponding centroid for each class in its space (Stage 1 of Fig. \ref{fig:arch}). This is used to generate pseudo-labels in the next stage.

\subsection{Stage 2: Pseudo-Labelling stage} \label{pseudo-label}
For the training of the target domain, pseudo-labels are generated. The inputs to the backbone are the target images, which yield a set of feature maps. To isolate the unknown classes, we use Local Outlier Factor (LOF) \cite{breunig2000lof}. The LOF tries to determine the deviation of the data points from local densities. If a point is not within the reachability space in any of the points, it will be considered an outlier. We propose an ablation in Section \ref{experiments} to determine the optimum feature map dimension for outlier detection and removal.

The outliers are assigned "unknown" class. The known classes representations are then assigned a class by querying the nearest class centroid, obtained from Section \ref{warmup}. Since the Open-Set label detection runs iteratively, we eliminate the need of highly complex outlier detection algorithms. As the representations become more discriminative, the separating ability of open-set classes increases. Such approaches have empirically shown promising results in "learning with noisy label" literature \cite{wang2018iterative}. To improve the accuracy of pseudo-labels, we propose re-computing them after every $K$ iterations of adaptation stage.

\subsection{Stage 3: Adaptation Stage} \label{adaptation-stage}
We propose the use of graph neural network with attentional aggregation. The GNN is a fully connected undirected graph, with nodes representing image and its domain features. There are two types of edges present in the graph. (a) Intra-edges $E_{self}$: connecting one source domain to other source domains and one target to another, (b) Inter-edges $E_{cross}$: connecting source and target domains together. Consider $^{(l)}\mathbf{x}^\mathcal{U}_i$, which represents image $i$ present in domain $\mathcal{U} \in \{S, T\}$ at an intermediate layer $l$. The message passing step can be represented as,
\begin{equation}
    ^{(l + 1)}\mathbf{x}_i^\mathcal{U} = {}^{(l)}\mathbf{x}_i^\mathcal{U} + \text{MLP}([^{(l)}\mathbf{x}_i^\mathcal{U}|\mathbf{m}_{E \to i}]) \\
\end{equation}
Where $\mathbf{m}_{E \to i}, E \in \{E_{self}, E_{cross}\}$ is a message pass, showing the aggregation of information from first order neighbours of the node, and $[.|.]$ denotes a concatenation operation. Following \cite{sarlin2020superglue}, we define a fixed number of layers $L$ (starting from $l = 1$), with $E = E_{self}$ if the layer is odd and $E = E_{cross}$ if the layer is even.

For attentional aggregation, self edges are used for self attention and cross edges for cross attention \cite{vaswani2017attention}. Given the node $^{(l)}\mathbf{x}^\mathcal{U}_i$, the key $\mathbf{k}_j$, value $\mathbf{v}_j$ and query $\mathbf{q}_i$ can be obtained as,
\begin{gather}
    \mathbf{q}_i = \mathbf{W}^{(l)}_1 \mathbf{x}^\mathcal{U}_i + \mathbf{b}_1^{(l)}
    \\
    \begin{bmatrix}
    \mathbf{k}_j \\ \mathbf{v}_j
    \end{bmatrix} = 
    \begin{bmatrix}
    \mathbf{W}_2 \\ \mathbf{W}_3
    \end{bmatrix}^{(l)}
    \mathbf{x}^\mathcal{U}_j + 
    \begin{bmatrix}
    \mathbf{b}_2 \\ \mathbf{b}_3
    \end{bmatrix}^{(l)}
\end{gather}
Where, $\mathbf{W}_{t}^{(l)}, \mathbf{b}_{t}^{(l)}; t \in \{1, 2, 3\}$ represent weight matrix and bias at layer $l$ to determine query, key and value respectively. This can be used to formulate the message as,
\begin{gather}
    \mathbf{m}_{E \to i} = \sum_{j: (i, j) \in E} \alpha_{ij}\mathbf{v}_j \\
    \alpha_{ij} = \text{Softmax}_j(\mathbf{q}_i^T\mathbf{k}_j)
\end{gather}
To improve the model expressivity, multi - head attention \cite{vaswani2017attention} can be used. In such case, the message passing expression can be shown as,
\begin{equation}
    \mathbf{m}_{E \to i} = \mathbf{W}^{(l)}(\mathbf{m}^1_{E \to i}|\mathbf{m}^2_{E \to i}|...|\mathbf{m}^h_{E \to i})
\end{equation}
Where $h$ denotes the number of attention heads and $\mathbf{W}^{(l)}$ denotes the linear transformation at layer $l$. Such a model is proposed to improve the data association between the source and target domain and at the same time, only match the essential characteristics between multiple domains.

\paragraph{Loss function} We intend to train the backbone for domain embedding and feature extractor in an offline fashion. The adaptation stage shall be trained online and used for inference. The source domain shall be trained using the available labels from the dataset, whereas the target domain will utilize the pseudo-labels obtained from Section \ref{pseudo-label}. The two components (source and target class loss) are used as the loss for online training and can be represented as,
\begin{equation}
    \mathcal{L} = \mathcal{L}^S_{CE} + \lambda \mathcal{L}^T_{CE}
\end{equation}
Where $\mathcal{L}^\mathcal{U}_{CE}$ is the cross-entropy loss for source and target heads, and $\lambda$ denotes the weightage of target loss.

\section{Experimental protocol} \label{experiments}

\subsection{Datasets} \label{datasets}
We will be testing our performance on popular domain adaptation benchmarks like Office31 \cite{DBLP:journals/corr/KoniuszTP16}, Office-Home \cite{DBLP:journals/corr/abs-1812-08974}, VisDA-2017 \cite{DBLP:journals/corr/abs-2003-13183} and DomainNet \cite{DBLP:journals/corr/abs-2008-11687}. For more details on the dataset composition, refer the supplementary material.

\subsection{Model architectures} \label{model-architectures}
For Stage 1, we propose offline training of domain embeddings as well as the backbone parameters. We intend to vary the backbone based on the datasets used for a fair comparison with other methods. For Office31 and Office-Home, we plan to use ResNet50 \cite{he2016deep}. For large-scale datasets such as VisDA and DomainNet, we will use ResNet101 as the backbone. The networks will be pre-trained on ImageNet, but will not be frozen. While the exact dimension of the feature vector will be decided based on ablation study in Section \ref{ablation-study}, we propose a dimension of $d = 1024$ for both the domain embedding and the bottleneck of feature extractor. 
The graph neural network with attentional aggregation will require nodes equal to the number of domains $m + n$, where $n$ and $m$ are the number of source and target domains, respectively. We intend to use $L = 9$ layers in the initial experimentation setup, with $h = 4$ heads in multi-head attention. For iterative refinement of target representation, we propose $K = 10,000$ for small scale datasets (Office31 and Office-Home) and $K = 50,000$ for large scale datasets (VisDA and DomainNet). We will also be doing a random hyperparameter search to find optimal values of hyperparameters.

\subsection{Experimental setting} \label{baseline}
The proposed method does not have any underlying assumptions regarding Domain Adaptation settings. Hence, the same model should be able to tackle problems such as \textit{\texttt{1}S\texttt{1}T} \cite{pmlr-v37-ganin15}, \textit{\texttt{1}SmT} \cite{Isobe_2021_CVPR}, \textit{nS\texttt{1}T} \cite{DBLP:journals/corr/abs-2002-12169} and \textit{nSmT} \cite{9242273}. We would be testing the same model under all the mentioned settings and compare it with their respective SOTA results as a baseline. All the mentioned experiments will be done for open set. The accuracy would be documented for each dataset, for different combinations of source and target domains. We would also analyse the behaviour of the proposed framework with the ratio of unknown class. All the experiments would be performed with 3 different seed values. The final results would be reported by taking average of all such accuracies. For information regarding performances of previous methods in these settings, refer the supplementary material.

We plan to implement the model using PyTorch, trained using Stochastic Gradient Descent (SGD) optimizer and having an initial learning rate $10^{-3} - 10^{-2}$ with cosine learning rate schedule.


\subsection{Ablation study} \label{ablation-study}
\subsubsection{Testing the effect of domain embedding} \label{d-embedding}
While most of the recent approaches to domain generalization focused on learning domain invariant features, in the proposed paper, we are investigating a domain adaptive approach \cite{dubey2021adaptive}. To test the effect of domain embedding on the model's performance, we will train two similar models, one with domain embedding attached to the feature vector and one without it. Then we will compare their performance to quantify the effect of domain embedding on model accuracy. Further, we will be testing different functions to add the domain embedding to the feature vector, namely, concatenation and element-wise multiplication.


\subsubsection{Attentional aggregation vs affinity matrix} \label{attention}
We used Attentional Aggregation to obtain a local representation of a node based on its first-order neighbours in the previous layer. But we can also use an \textit{Affinity  matrix} \cite{roy2021curriculum, luo2020progressive} to represent the semantic similarity \cite{kipf2016semi} between the images. We will be comparing the effect of both representations on the overall accuracy of the model.

\subsubsection{Testing outlier prediction} \label{lof}
 The proposed method uses centroid space to define pseudo-labels for objects. We are using LOF \cite{wang2018iterative} to extract outliers which will be used to assign the \textit{unknown} class. We intend to check how well LOF handles outliers with varying input dimension, for this test we plan to change the nodes in the output layer of the network before LOF, effectively changing the dimensions of input to LOF and measure its effect on the overall accuracy, finding out the optimal dimension. We will also be tracking the improvement of pseudo-label accuracy per completion of an iteration. We predict that while the initial pseudo-label accuracy would be low, it should continually increase as training progresses. Hence, pseudo-label vs iteration curve should be non-decreasing.
 
\section{Limitations and conclusion} \label{discussion-conclusion}
We address a potential issue that can arise in our approach. This involves the use of LoF, requiring outliers to isolate them from the main data. Our model assumes sufficient number of unknown classes to be classified as an outlier. This might restrict the number of unknown classes our model can handle. 

In this work, we propose a novel domain adaptation setting 'Open-Set Multi-Source Multi-Target' and a plausible solution for the same. Specifically, we leverage graph neural networks for message passing amongst different domains, along with domain embedding for providing higher level information to each sample. An extension to the proposed approach is the generalization ability on unseen domains which can be explored in future works.




\bibliographystyle{unsrt}
\bibliography{ref.bib}

\end{document}


\section{Proposed Algorithm}
\RestyleAlgo{ruled}
\SetKwComment{Comment}{// }{}

\begin{algorithm}[H]
\caption{DEGAA: Offline Training Paradigm}\label{alg:one}
\textbf{Data. }$S = \cup_{i = 1}^{n}S_i, T = \cup_{j = 1}^{m}T_j, D = S \cup T$\\
\underline{\textbf{Step 1:} \textit{Domain Embedding Extraction}}\\
\textbf{Input. } Number of sampled domains $N_t$, Number of support points $N_s$ and query points $N_q$ \\
\textbf{Initialize.} Feature Extractor $G(\mathbf{x}; \psi)$ initialized with pre-trained ImageNet weights\\
\For{$\text{t = 1 \textrm{To} N}$}{
  $D_t \gets \textsc{Randomly Sample}(D, N_t)$ \Comment*[r]{sample $N_t$ domains}
  \For{$d$ in $D_t$}{
    $S_d, S_q \gets \textsc{Randomly Sample}(D_d, N_s), \textsc{Randomly Sample}(D_d, N_q)$ \;
  }
  $\mathbf{\hat{\mathbf{\mu}}}_{D_t} \gets \textsc{Kme}(S_d, G(\mathbf{x}; \psi))$ \;
  $J_{\psi} \gets \textsc{Prototypical Loss}(\mathbf{\hat{\mathbf{\mu}}}_{D_t}, S_t, G(\mathbf{x}; \psi))$ \Comment*[r]{Following \cite{dubey2021adaptive}}

  $\psi \gets \textsc{Sgd}(J(t), \psi)$ \;
}
\textbf{Output. }$d_e \gets \textsc{Kme}(D, G(\mathbf{x}; \psi))$ \;
\underline{\textbf{Step 2:} \textit{Warm - Up}}\\
\textbf{Input. } Domain Embedding $d_e$, source images and labels $(\mathbf{x}^{S_i}_{j}, y^{S_i}_{j})_{j = 1}^{p_i} \in (S_i)_{i = 1}^{n}$, training images $(\mathbf{x}^{\widehat{S}}_{j}, y^{\widehat{S}}_{j})_{j = 1}^{n'} \in \widehat{S} \subset S$ \\
\textbf{Initialize.} Feature Extractor $F(\mathbf{x}; \psi)$ initialized with pre-trained ImageNet weights \\
\For{t = 1 To N'}{
   $J_{\theta} \gets \textsc{Cross Entropy}(y^{\widehat{S}}_{j}, F(\textsc{Concat}(\mathbf{x}^{\widehat{S}}_{j}, d_e)))$ \Comment*[r]{Supervised training}
   $\theta \gets \textsc{SGD}(J(t), \theta)$ \;
}
\underline{\textbf{Step 3:} \textit{Compute Centroids}}\\
\For{$\text{i = 1 \textrm{To} n}$}{
  \For{$\text{j = 1 \textrm{To} }p_i$}{
    $\mathcal{T} \gets F(\textsc{Concat}(\mathbf{x}^{S_i}_{j}, d_e); \theta)$ \;
  }
}
$\mathcal{C} \gets \textsc{Centroids}(\mathcal{T})$ \Comment*[r]{Per - class centroid from source feature maps}
\underline{\textbf{Step 4:} \textit{Pseudo - Labelling and Adaptation Stage}}\\
\textbf{Input. } Domain Embedding $d_e$, trained backbone $F(\mathbf{x};\theta)$, number of episodes per batch $K$, batch for $K$ episodes $(\mathbf{x}^{S_K}_{i}, y^{S_K}_{i})_{i = 1}^{n'} \in S_K \subset S, (\mathbf{x}^{T_K}_{j})_{j = 1}^{m'} \in T_K \subset T$, target loss weight $\lambda$\\
\For{t = 1 To N''}{
$S_K, T_K \gets \textsc{Randomly Sample}(S, n'), \textsc{Randomly Sample}(T, m')$ \;
\For{k = 1 To K}{
  \For{j = 1 To m'}{
    $D'_t \gets F(\textsc{Concat}(\mathbf{x}^{T_K}_{j}, d_e); \theta)$ \Comment*[r]{Concatenated feature maps(target)
    } 
  }
  $D'_k, D'_u \gets \textsc{Lof}(D'_t)$ \Comment*[r]{Known classes $D'_k$ and unknown $D'_u$}
  $Y_{pseudo} \gets \textsc{Knn}(D'_k, \mathcal{C})$ \Comment*[r]{Assign nearest centroid class}
  \For{i = 1 To n'}{
    $D'_s \gets F(\textsc{Concat}(\mathbf{x}^{S_K}_{i}, d_e); \theta)$ \Comment*[r]{Concatenated feature maps(source)}
    $Y_s \gets y^{S_K}_{i}$ \;
  }
  $\widehat{Y}_s, \widehat{Y}_t \gets \textsc{Softmax}(\textsc{Gaa}(D'_s, D'_k))$ \Comment*[r]{Source and target labels using GAA}
  $J_{\theta} \gets \textsc{Cross Entropy}(\widehat{Y}_s, Y_s) + \lambda\textsc{Cross Entropy}(\widehat{Y}_{t}, Y_{pseudo})$ \;
  $\theta \gets \textsc{SGD}(J(t), \theta)$ \;
}
}
\end{algorithm}


\section{Dataset Details} \label{datasets}
\begin{itemize}[noitemsep,nolistsep,leftmargin=*]
    \item \textbf{Office31:} The Office31 dataset \cite{DBLP:journals/corr/KoniuszTP16} contains 31 object categories in three domains: Amazon, DSLR and Webcam. The object categories include everyday objects such as keyboards, laptops and file cabinets.
    Amazon Domain contains 2817 images captured against clean background and at a unified scale. The DSLR domain contains 498 high resolutions images while the WebCam contains 795 low resolutions images.
    \item \textbf{OfficeHome:} The OfficeHome dataset \cite{DBLP:journals/corr/abs-1812-08974} contains images in 4 different domains consisting of 65 object categories found typically in Office and Home settings. Total 15,500 images are present with images in each class vary between 70 images to a maximum of 99 images.
    \item \textbf{VisDA-2017:} This large scale dataset \cite{DBLP:journals/corr/abs-2003-13183} contains over 280,000 images across 12 categories. The training images are generated from the same object under different conditions while the validations images are sourced from MSCOCO.
    \item \textbf{DomainNet:} The DomainNet dataset \cite{DBLP:journals/corr/abs-2008-11687} contains over half a million images in 6 different domains, each consisting of 345 categories of objects. The domains include clipart, real world photos, sketches, infograph, QuickDraw and paintings

\end{itemize}


\section{Tables}

Following previous works \textbf{OS} indicates normalized accuracy for all the classes including the unknown as one class and \textbf{OS*} shows normalized accuracy only on known classes.

\begin{table}[H]
\caption{Classification Accuracy (\%) of open set domain adaptation tasks on Office-31 (ResNet-50)}
\label{Office31}
\centering
\resizebox{\columnwidth}{!}{\begin{tabular}{lcccccccccccccc} 
\toprule
\multirow{2}{*}{Method} & \multicolumn{2}{c}{\textbf{A}$\rightarrow$\textbf{W}} & \multicolumn{2}{c}{\textbf{D}$\rightarrow$\textbf{W}} & \multicolumn{2}{c}{\textbf{W}$\rightarrow$\textbf{D}} & \multicolumn{2}{c}{\textbf{A}$\rightarrow$\textbf{D}} & \multicolumn{2}{c}{\textbf{D}$\rightarrow$\textbf{A}} & \multicolumn{2}{c}{\textbf{W}$\rightarrow$\textbf{A}} & \multicolumn{2}{c}{\textbf{Avg}} \\ 
\cmidrule(r){2-3}
\cmidrule(r){4-5}
\cmidrule(r){6-7}
\cmidrule(r){8-9}
\cmidrule(r){10-11}
\cmidrule(r){12-13}
\cmidrule(r){14-15}
 & OS & OS$^*$ & OS & OS$^*$ & OS & OS$^*$ & OS & OS$^*$ & OS & OS$^*$ & OS & OS$^*$ & OS & OS$^*$ \\
\midrule
ResNet \cite{he2016deep} & 82.5 & 82.7 & 85.2 & 85.5 & 94.1 & 94.3 & 96.6 & 97.0 & 71.6 & 71.5 & 75.5 & 75.2 & 84.2 & 84.4 \\
ATI-$\lambda$ \cite{Busto_2017_ICCV} & 87.4 & 88.9 & 84.3 & 86.6 & 93.6 & 95.3 & 96.5 & 98.7 & 78.0 & 79.6 & 80.4 & 81.4 & 86.7 & 88.4 \\
OSBP \cite{Saito_2018_ECCV} & 86.5 & 87.6 & 88.6 & 89.2 & 97.0 & 96.5 & 97.9 & 98.7 & 88.9 & 90.6 & 85.8 & 84.9 & 90.8 & 91.3 \\
STA \cite{Liu_2019_CVPR} & 89.5 & 92.1 & 93.7 & 96.1 & 97.5 & 96.5 & 99.5 & 99.6 & 89.1 & 93.5 & 87.9 & 87.4 & 92.9 & 94.1 \\
JPOT \cite{ijcai2020-352} & 92.8 & 92.2 & 95.2 & 96.0 & 98.1 & 96.2 & 99.5 & 98.6 & 93.0 & 94.1 & 88.9 & 88.4 & 94.6 & 94.3 \\
\midrule
Ours & \multicolumn{14}{c}{-} \\
\bottomrule
\end{tabular}}
\end{table}

\begin{table}[H]
\caption{Classification accuracy (\%) of open set domain adaptation tasks on Office-Home (ResNet-50)}
\label{OfficeHome}
\centering
\resizebox{\columnwidth}{!}{\begin{tabular}{lccccccccccccc}
\toprule
Method & \textbf{Ar$\rightarrow$Cl} & \textbf{Pr$\rightarrow$Cl} & \textbf{Rw$\rightarrow$Cl} & \textbf{Ar$\rightarrow$Pr} & \textbf{Cl$\rightarrow$Pr} & \textbf{Rw$\rightarrow$Pr} & \textbf{Cl$\rightarrow$Ar} & \textbf{Pr$\rightarrow$Ar} & \textbf{Rw$\rightarrow$Ar} & \textbf{Ar$\rightarrow$Rw} & \textbf{Cl$\rightarrow$Rw} & \textbf{Pr$\rightarrow$Rw} & \textbf{Avg.}\\
\midrule
ResNet \cite{he2016deep} & 53.4 & 52.7 & 51.9 & 69.3 & 61.8 & 74.1 & 61.4 & 64.0 & 70.0 & 78.7 & 71.0 & 74.9 & 65.3 \\
ATI-$\lambda$ \cite{Busto_2017_ICCV} & 55.2 & 52.6 & 53.5 & 69.1 & 63.5 & 74.1 & 61.7 & 64.5 & 70.7 & 79.2 & 72.9 & 75.8 & 66.1 \\
OSBP \cite{Saito_2018_ECCV} &  56.7 & 51.5 & 49.2 & 67.5 & 65.5 & 74.0 & 62.5 & 64.8 & 69.3 & 80.6 & 74.7 & 71.5 & 65.7 \\
STA \cite{Liu_2019_CVPR} & 58.1 & 53.1 & 54.4 & 71.6 & 69.3 & 81.9 & 63.4 & 65.2 & 74.9 & 85.0 & 75.8 & 80.8 & 69.5 \\
JPOT \cite{ijcai2020-352} &  59.6 & 54.2 & 54.6 & 72.3 & 70.1 & 82.1 & 62.9 & 68.3 & 75.1 & 84.8 & 77.4 & 81.2 & 70.2 \\
PGL \cite{luo2020progressive} & 61.6 & 58.4 & 65.0 & 77.1 & 72.0 & 83.0 & 68.8 & 72.2 & 78.6 & 85.9 & 82.8 & 82.6 & 74.0 \\
\midrule
Ours & \multicolumn{13}{c}{-} \\
\bottomrule
\end{tabular}}
\end{table}

\begin{table}[H]
\caption{Classification accuracy (\%) of open set domain adaptation tasks on VisDA-2017 (VGGNet)}
\label{visda17}
\centering
\begin{tabular}{lccccccccc}
\toprule
Method & Bic & Bus & Car & Mot & Tra & Tru & UNK & OS & OS$^*$\\
\midrule 
AATI-$\lambda$ \cite{Busto_2017_ICCV} & 46.2 & 57.5 & 56.9 & 79.1 & 81.6 & 32.7 & 65.0 & 59.9 & 59.0 \\
OSBP \cite{Saito_2018_ECCV} & 51.1 & 67.1 & 42.8 & 84.2 & 81.8 & 28.0 & 85.1 & 62.9 & 59.2\\
STA \cite{Liu_2019_CVPR} & 52.4 & 69.6 & 59.9 & 87.8 & 86.5 & 27.2 & 84.1 & 66.8 & 63.9\\
PGL \cite{luo2020progressive} & 93.5 & 93.8 & 75.7 & 98.8 & 96.2 & 38.5 & 68.6 & 80.7 & 82.8 \\
\midrule
Ours & \multicolumn{9}{c}{-} \\
\bottomrule
\end{tabular}
\end{table}

\begin{table}[H]
\caption{Classification accuracy (\%) of Multi Source Open Set domain adaptation tasks on Office-31.}
\label{OSMSDA}
\centering
\begin{tabular}{lcccccccc}
\toprule
\multirow{2}{*}{Method} & \multicolumn{2}{c}{\textbf{AD}$\rightarrow$\textbf{W}} & \multicolumn{2}{c}{\textbf{AW}$\rightarrow$\textbf{D}} & \multicolumn{2}{c}{\textbf{WD}$\rightarrow$\textbf{A}} & \multicolumn{2}{c}{\textbf{Avg}} \\ 
\cmidrule(r){2-3}
\cmidrule(r){4-5}
\cmidrule(r){6-7}
\cmidrule(r){8-9}
 & OS & OS$^*$ & OS & OS$^*$ & OS & OS$^*$ & OS & OS$^*$ \\
\midrule
MOSDANET \cite{Rakshit2020MultisourceOD} & 99.0 & 98.2 & 99.4 & 98.3 & 81.0 & 79.3 & 93.1 & 91.9 \\
\midrule
Ours & \multicolumn{8}{c}{-} \\
\bottomrule
\end{tabular}
\end{table}

\begin{table}[H]
\caption{Comparison with the state-of-the-art methods on the DomainNet dataset.}
\label{DomainNet}
\centering
\begin{tabular}{lccccccccc}
\toprule
Method & R$\rightarrow$S & R$\rightarrow$C & R$\rightarrow$I & R$\rightarrow$P & P$\rightarrow$S & P$\rightarrow$R & P$\rightarrow$C & P$\rightarrow$I & \textbf{Avg} \\
\midrule
CGCT \cite{roy2021curriculum} & 48.9 & 60.3 & 26.9 & 57.1 & 43.4 & 58.8 & 48.5 & 21.7 & 45.7 \\
D-CGCT \cite{roy2021curriculum} & 48.4 & 59.6 & 25.3 & 55.6 & 45.3 & 58.2 & 51.0 & 21.7 & 45.6 \\
DCC \cite{Li_2021_CVPR} & 43.1 & - & - & 50.25 & 43.66 & 56.90 & - & - & 48.5 \\
\midrule
Ours & \multicolumn{9}{c}{-} \\
\bottomrule
\end{tabular}
\end{table}

\bibliographystyle{unsrt}
\bibliography{ref.bib}